\documentclass[a4paper, 10pt, conference]{ieeeconf}      

\IEEEoverridecommandlockouts                              
\usepackage{graphicx} 
\usepackage{amsmath} 
\usepackage{amssymb}  
\usepackage{algorithm}
\usepackage{algorithmic}
\usepackage{xcolor}
\usepackage{soul}

\makeatletter
\let\NAT@parse\undefined
\makeatother
\usepackage{hyperref}
\usepackage{float}
\graphicspath{{figures/}}

\title{\LARGE \bf A Road-map to Robot Task Execution with the Functional Object-Oriented Network}

\author{David Paulius, Alejandro Agostini, Yu Sun, and Dongheui Lee\\
\thanks{
David Paulius, Alejandro Agostini, and Dongheui Lee are members of the Human-centered Assistive Robotics (HCR) group at the Technical University of Munich, Germany.
Yu Sun leads the Robot Perception and Action Lab (RPAL) at the University of South Florida, Tampa, FL, USA.
\newline(\textit{Corresponding Author}: David Paulius --  \tt{david.paulius@tum.de)}}%
}

\begin{document}

\maketitle

\thispagestyle{empty}
\pagestyle{empty}

\begin{abstract}
Following work on joint object-action representations, the functional object-oriented network (FOON) was introduced as a knowledge graph representation for robots.
Taking the form of a bipartite graph, a FOON contains symbolic or high-level information that would be pertinent to a robot's understanding of its environment and tasks in a way that mirrors human understanding of actions.
In this work, we outline a road-map for future development of FOON and its application in robotic systems for task planning as well as knowledge acquisition from demonstration.
We propose preliminary ideas to show how a FOON can be created in a real-world scenario with a robot and human teacher in a way that can jointly augment existing knowledge in a FOON and teach a robot the skills it needs to replicate the demonstrated actions and solve a given manipulation problem.
\end{abstract}

\section{Introduction}
An ongoing trend for research in robotics is the development of robots that can jointly understand human intention and action and execute manipulations for human domains.
A key component for such intelligent and autonomous robots is a knowledge representation~\cite{paulius2019survey}, which would allow robots to understand its actions in a way that mirrors human understanding of action and affordance~\cite{Gibson_1977}.
Prior to this work, we introduced the functional object-oriented network (FOON) as a knowledge representation for service robots~\cite{paulius2016functional}, which was inspired by previous work on joint object-action representation~\cite{SunRAS2013,Lin2015a}.
A FOON innately describes the relationship between objects and manipulation actions as nodes. 
Ideally, graphs are formed from demonstrations of action, and they can be combined into a single network from which knowledge can be retrieved as task sequences~\cite{paulius2016functional}. 
We also showed how existing knowledge can be used to learn ``new'' concepts based on semantic similarity~\cite{paulius2018functional}, and in~\cite{paulius2021weighted}, we introduced a new retrieval algorithm that considers a robot's capabilities to find an optimal task sequence.

\begin{figure}[t]
	\centering
	\includegraphics[trim={0.25cm 0cm 0.25cm 0cm},clip,width=7.5cm]{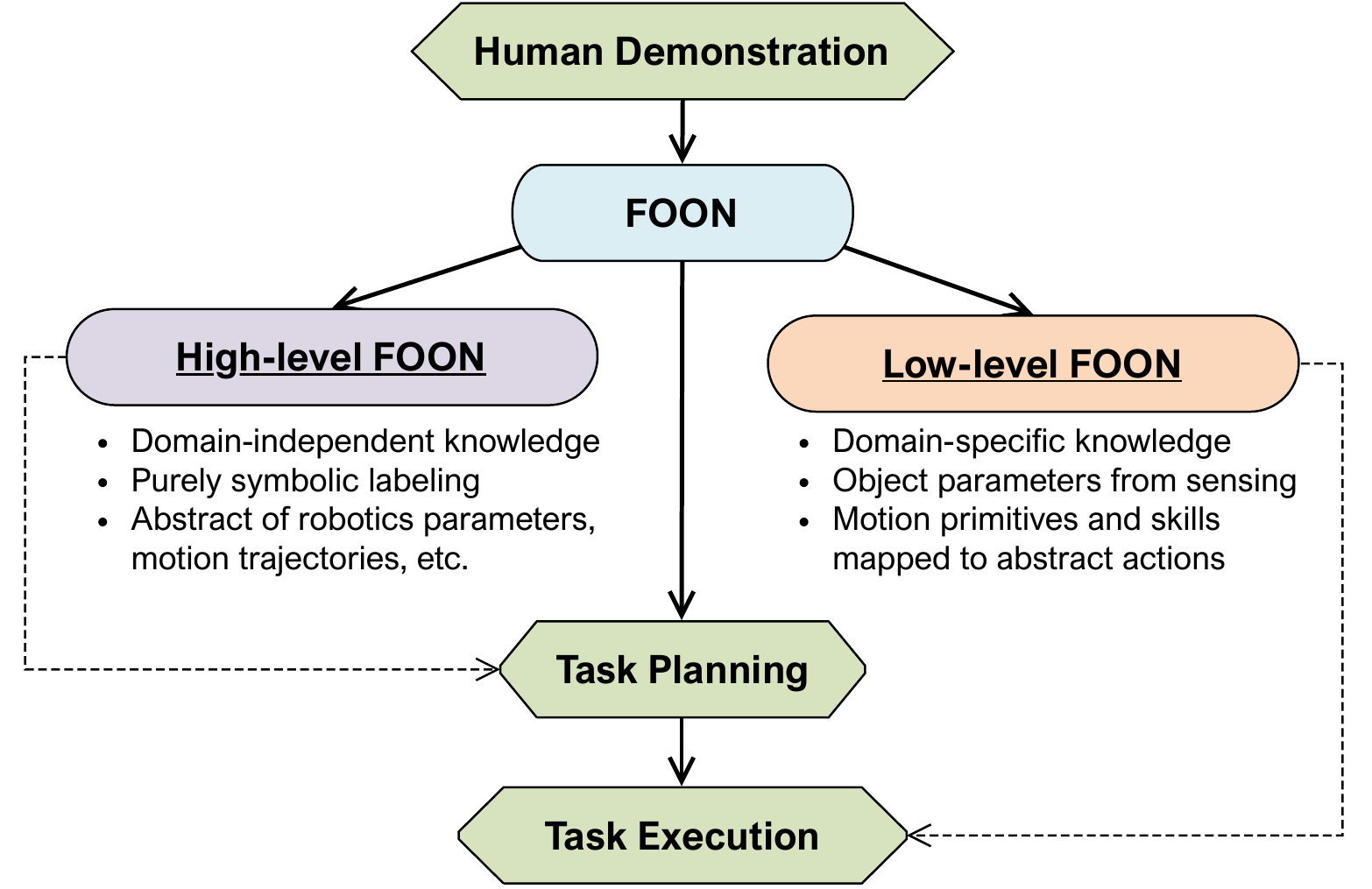}
	\caption{Overview of the proposed pipeline incorporating elements of LfD with FOON knowledge creation and retrieval. 
	}
	\label{fig:overview}
\end{figure}

In this preliminary work, we propose a road-map for further applications of FOON and to integrate it in task planning and execution in order to take advantage of the richness of such knowledge graphs.
Prior to this paper, little has been done to incorporate FOON into a robotic system, as the knowledge in a FOON is too abstract for immediate use in robotic manipulation planning~\cite{agostini2020manipulation}.
Furthermore, the FOON dataset is comprised of subgraphs whose contents were collected from manual annotation by human volunteers.
To address these limitations, we propose a pipeline (shown as Figure~\ref{fig:overview}) that integrates learning from demonstration (LfD) such that: 1) we can construct graphs directly from observation, and 2) we can integrate symbolic knowledge in FOON with physical aspects, such as a robot's motion primitives, object parameters, and trajectories.
Although some work has been done to semi-automatically construct graphs from videos~\cite{jelodar2018long}, there is merit to learning in a teacher-student setting, where a human demonstrator can teach necessary skills to a robot to augment knowledge gathered from videos.

This paper is organized as follows: in Section \ref{sec:FOON}, we give a short overview of the FOON structure and algorithms in a FOON terminology.
In Section \ref{sec:lfd}, we introduce our preliminary ideas for the integration of LfD in a learning setting to ground an abstract, \textit{domain-independent} FOON into a \textit{domain-specific} FOON.
Finally, in Section \ref{sec:con}, we summarize our ideas and give an overview of future work.

\section{Functional Object-Oriented Network}
\label{sec:FOON}

\subsection{Basics of a FOON}
A FOON consists of two types of nodes in its bipartite structure: \textit{object nodes} and \textit{motion nodes}.
Object nodes refer to objects that are used in activities, including tools, utensils, ingredients or components, while motion nodes refer to actions that can be performed on said objects.
Presently, we use FOON to represent activities in cooking.
Objects are identified by their object type, its states, and, in some cases, its make-up of ingredients or components; motions are identified by a motion or action type, which can refer to a manipulation (e.g., pouring, cutting, and shaking) or a non-manipulation action (e.g., frying or baking).
As a result of executing actions, objects may take on new states or conditions; state transitions are conveyed in \textit{functional units}, which are collections of object nodes and a motion node before and after an action takes place.
Therefore, when using FOON, it is important for a robot to identify states to determine when an action or goal has been completed.

\begin{figure}[t]
	\centering
    \includegraphics[width=7cm]{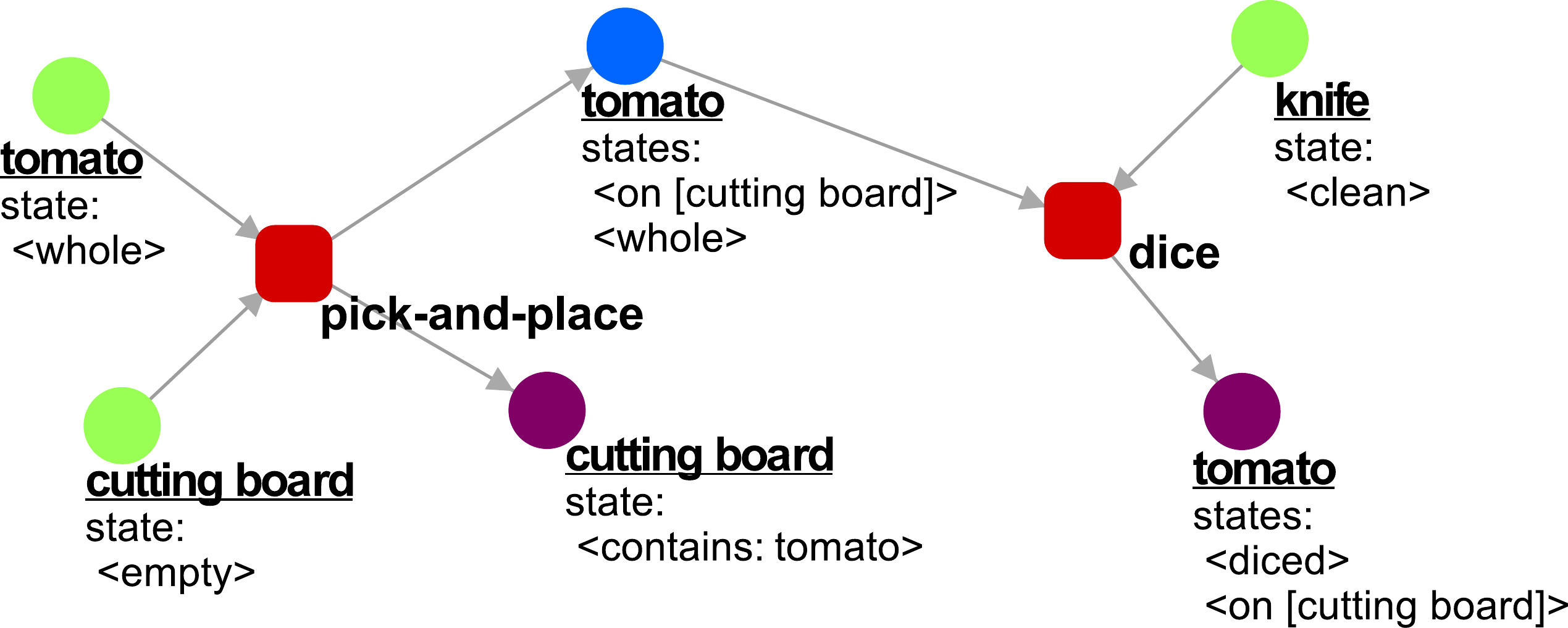}
	\caption{An example of two functional units, which describe placing a tomato on a cutting board and dicing it with a knife (best viewed in colour).
	Object nodes are denoted by circles, while motion nodes are denoted by squares.
	Here, input-only nodes are depicted in green, output-only nodes are depicted in purple, and nodes that are both input and output are depicted in blue.
	}
	\label{fig:unit}
\end{figure}

Figure \ref{fig:unit} illustrates an example of two functional units describing the action of placing a tomato on a cutting board and dicing it with a knife. 
Without considering states, we have the objects \textit{cutting board}, \textit{tomato}, and \textit{knife}.
Note, however, that there are several instances of these objects, as their states will change as a result of execution; this is analogous to Petri Nets~\cite{Petri:2008}, where the firing of transitions cause a change in input place nodes.
Each functional unit has a motion node for picking and placing the tomato on the cutting board (\textit{pick-and-place}) and dicing the tomato (\textit{dice}).



\subsection{Creating a Universal FOON}
A FOON is typically created using annotations of activities as video demonstrations or, as we plan to explore in the near future, demonstrations from a human teacher.
When annotating, it is important to note the objects, actions, and state changes that have been observed and that are required to achieve a specific goal, such as preparing a meal or recipe.
As a result of this process, one can obtain a FOON \textit{subgraph} for several activities or recipes, which describes a sequence of functional units that capture information on the objects, manipulations and actions required to fulfill the task's goal.
The combination of two or more subgraphs form a \textit{universal} FOON.
This merging procedure consolidates all instances of object nodes and removes duplicate functional units that are common across different subgraphs or recipes~\cite{paulius2016functional}.
Presently, the FOON dataset provides 111 subgraph annotations from which a universal FOON has been created; these annotations along with helper code can be found online\footnote{FOONets (FOON Website) -- \url{http://www.foonets.com}}.

\subsection{Task Planning with FOON}
Aside from representing knowledge in a symbolic manner, a FOON can be used for problem solving through task planning.
Such a problem would entail answering the question: given a set of objects (e.g., tools, ingredients, or appliances), how can a robot create an object denoted as a node in a FOON?
Given that a robot can understand its environment, where it can ground instances of objects to nodes in FOON, we can rely on FOON to determine how those objects can be utilized to solve more complex problems.
This knowledge retrieval procedure is denoted as \textit{task tree retrieval}~\cite{paulius2016functional}, where a \textit{task tree} is a subgraph that describes the steps for solving a given problem based on the state of the robot's environment.

The principle behind the task tree retrieval algorithm combines ideas from breadth-first and depth-first search (BFS and DFS), where, starting from the goal node or sub-goal nodes, we perform a backward search to find the functional units needed to make them (DFS), and for each functional unit, we evaluate if their input conditions are met (BFS).
As such, this will require some knowledge of the initial set of objects that could be made or that are already available to the robot.
If we want to perform a task tree retrieval operation without immediately considering the availability of objects or to find the best course of action that a robot can successfully execute, then one can also consider building a \textit{path tree}~\cite{paulius2021weighted}, where we can retrieve possible combinations of action sequences that achieves the final goal.
This form of retrieval was introduced to find the optimal task tree based on the robot's success rate of executing its motion primitives.

\section{Learning from Demonstration for FOON}
\label{sec:lfd}

\begin{figure*}[t]
	\centering
	\includegraphics[trim={0cm .35cm 0cm 0cm},clip,width=13cm]{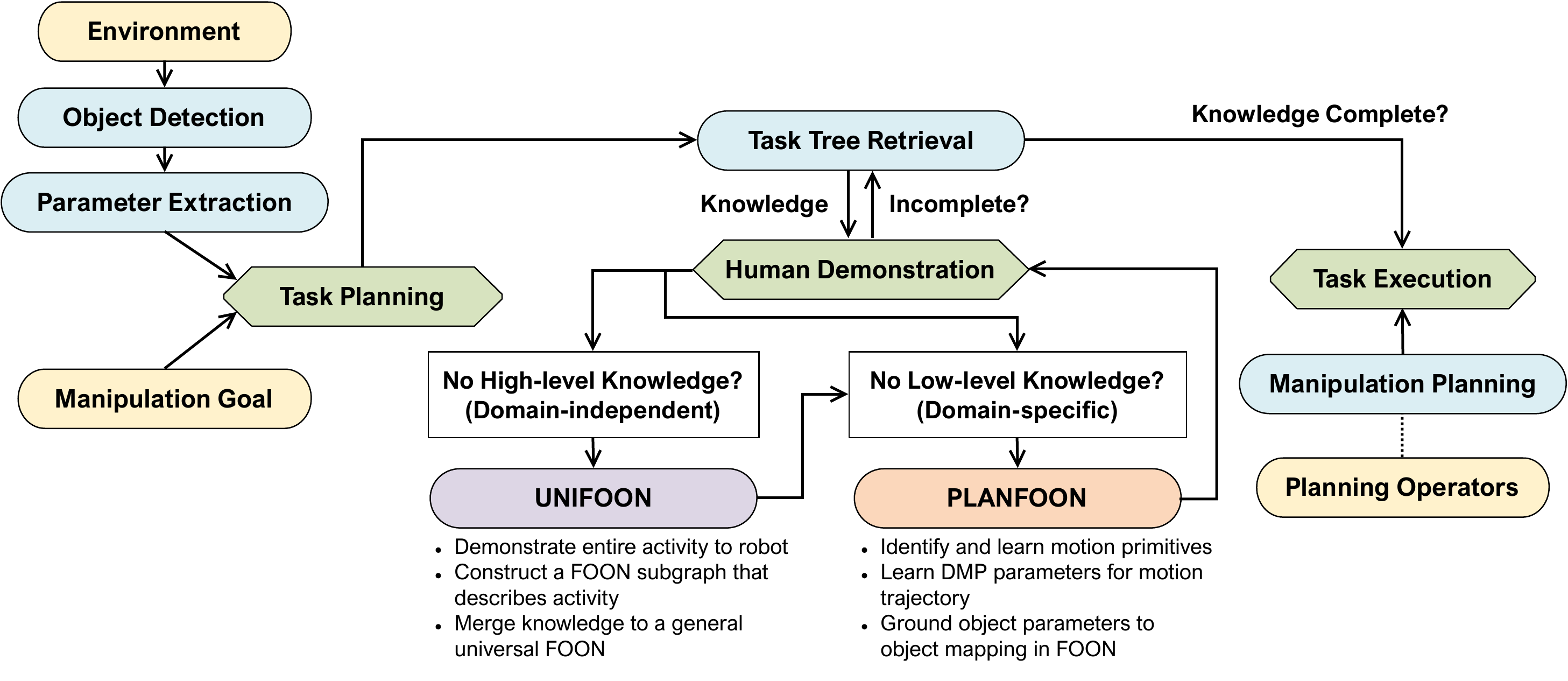}
 	\caption{Overview of our learning framework that integrates a domain-independent FOON ({\sc{unifoon}}) with a domain-specific FOON ({\sc{planfoon}}).}
	\label{fig:pipeline}
\end{figure*}

A FOON can be created using annotations from demonstrations.
However, up to this point, annotation was done manually, although semi-automatic annotation has been explored from video by using existing FOON knowledge to annotate never-before-seen videos~\cite{jelodar2018long}.
The problem introduced in this work entails annotation directly from human demonstration, where a robot will be shown a sequence of actions, which is performed by a teacher, at first hand.

In detail, the objective of this learning from demonstration pipeline is to connect a \textit{domain-independent} FOON to a \textit{domain-specific} representation, where a system can connect abstracted concepts in a universal FOON to the physical world and to relevant robotic properties.
This would require key components~\cite{paulius2019survey}, including object detection, grounding object instances to FOON nodes, action recognition, and the demonstration of key skills or motion primitives to perform actions represented as functional units.
In this paper, we refer to a domain-independent FOON as a universal FOON (or {\sc{unifoon}}) and a domain-specific FOON as a planetary FOON (or {\sc{planfoon}}).
We illustrate an overview of our framework as Figure \ref{fig:pipeline}.

\subsection{Identifying Objects for FOON}

Object nodes in a FOON relate to objects that are directly or indirectly manipulated for tasks.
In a regular FOON (or {\sc{unifoon}}), object nodes are typically described by their object type, state(s), and their ingredient composition. 
At the {\sc{planfoon}} level, our proposed learning framework would need to connect low-level information (based on signals and sensors) to the high-level attributes in {\sc{unifoon}}.
From a low-level perspective, the objective is to identify key features that can be used by a robotic system to identify the objects and their observed states that exist within its environment and that are used in a certain action segment.
Useful parameters, such as object poses, images, point clouds or models, object centroids and bounding boxes, can be extracted from a scene, and from these parameters, a robot can then ground object instances to the symbols found in FOON.
At the same time, from a high-level perspective, a robot should understand that these detected objects translate to their respective object node symbols in a FOON based on such parameters.

To achieve this, a learning task would comprise of training a model to output object and state labels after being fed these parameters as input. 
Such a model or system would need to perform two variations of classification: \textit{object classification} and \textit{state recognition}.
For object classification, a model can be used to learn how to detect different types of objects as used in a FOON regardless of their observed state, either in a supervised or unsupervised manner.
The more challenging problem lies in state recognition, where it is important to discriminate between different states, requiring lots of training data and thus more complex state-of-the-art approaches such as~\cite{jelodar2019joint}.
To ease the burden on the training procedure, we will first explore how we can simplify object and state detection by manually identifying parameter thresholds for a limited object-state sample size and then further refine them when it comes to different object instances.

\subsection{Learning Motion Primitives and Skills}
A FOON's motion nodes and functional units correspond to different types of actions, which usually involve interaction or manipulation of objects, such as tools, containers, ingredients, or appliances.
A functional unit can be equated to planning operators, which are often times defined using planning domain languages such as PDDL~\cite{mcdermott1998pddl}.
Using a graphical representation like FOON can allow for faster knowledge retrieval due to discretization of the problem domain. 
When demonstrating an action, it is important to learn trajectories as skills, which can be retained for replication.
One method of learning and representing trajectories is to use dynamic movement primitives (DMP)~\cite{ijspeert2013dynamical}.
Based on a start and end position, a DMP can be used to approximate the original demonstrated trajectory; forcing terms are included that will retain details of the trajectory, where the more forcing terms (as weights) are used, the closer the learned trajectory will be to the original demonstration.
These parameters as well as object centroids will be acquired from the object parameter extraction phase discussed earlier.

\subsection{Integrating Domain-specific Knowledge into FOON}
Combined with observable object parameters, skills as DMPs can be learned for planning operators represented in FOON, and they can be mapped to motion nodes of each functional unit in a {\sc{unifoon}}.
However, many of the actions that are present in a typical FOON are still too abstract to represent as a single motor skill or primitive.
For instance, let us consider the functional units in Figure \ref{fig:unit}. 
In the case of the picking-and-placing unit, this action can be decomposed into the sub-actions of: moving the robot's gripper to the tomato object, grasping the tomato, translating the gripper to the target location of the cutting board, and then releasing the object to complete the place.
In the case of the dicing unit, this action can be decomposed into the sub-actions of: moving the gripper to the knife, grasping it, moving the gripper with the knife to the tomato, executing the dice action, and returning the knife to its original location.
Therefore, the notion of actions in {\sc{unifoon}} cannot suffice, and the {\sc{planfoon}} level will have to provide a similar chaining of atomic skills to the abstract action. 

\begin{figure}[t]
    \centering
    \includegraphics[width=\columnwidth]{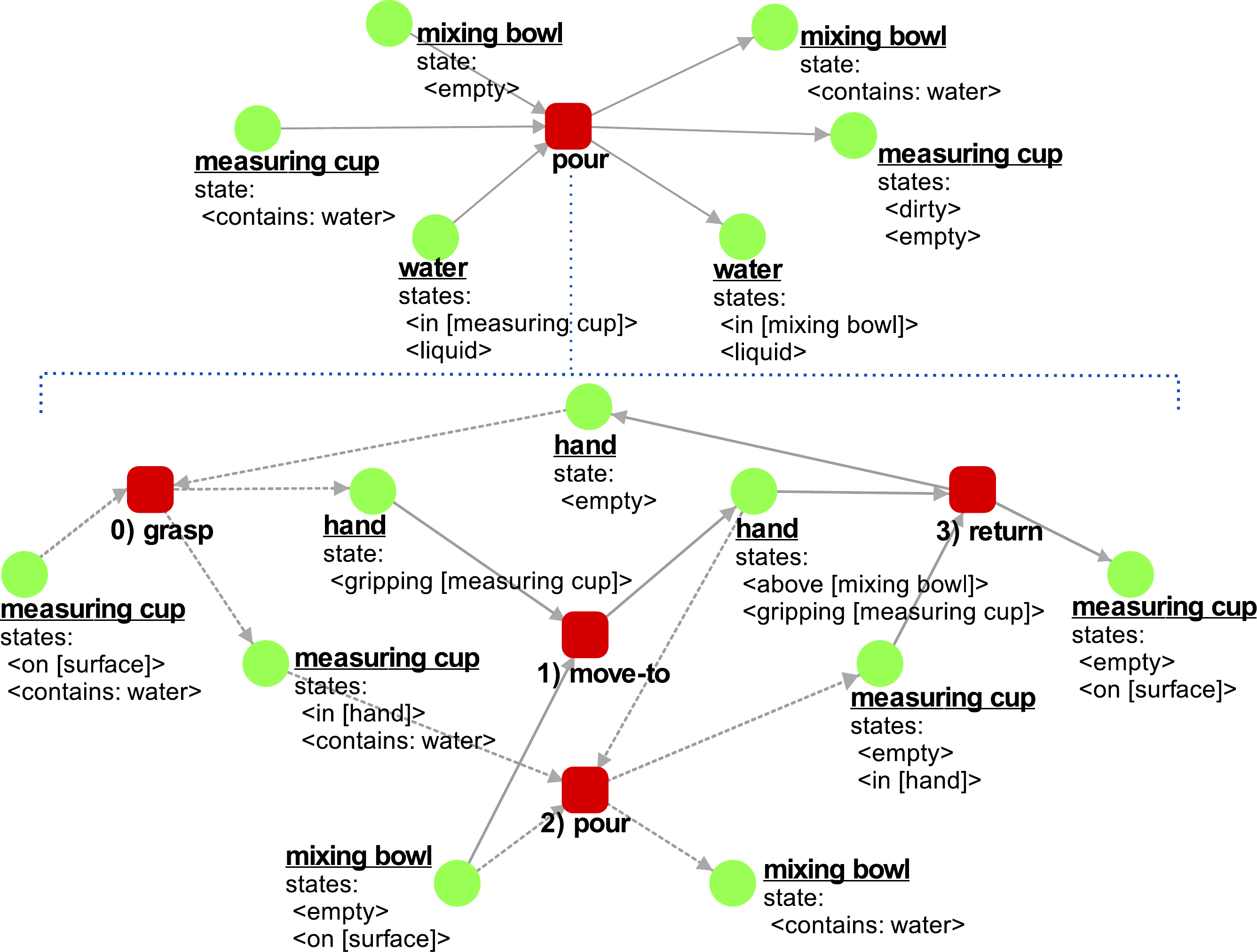}
    \caption{An example of a {\sc{unifoon}} to {\sc{planfoon}} mapping for the action of pouring water from a measuring cup to a bowl.
    At the {\sc{planfoon}} level, atomic motion primitives will map to {\sc{unifoon}} level actions.
    {\sc{planfoon}} actions will be connected to planning operators and DMPs.}
    \label{fig:integration}
\end{figure}

\subsubsection{Details}
We propose to introduce {\sc{planfoon}} as an additional hierarchy that will focus on the domain-specific details that are needed to physically execute the actions given by a {\sc{unifoon}}.
This notion of hierarchy is innately different to that in~\cite{paulius2018functional}, where the latter is perhaps better described as levels of abstraction.
A task planning problem would require searching for a task sequence first in the upper {\sc{unifoon}} level, which is then followed by the retrieval of execution-level details in the lower {\sc{planfoon}} level.
This allows us to take advantage of the rich and descriptive form of knowledge that is available from explicitly grounding to FOON.

Using learning from demonstration (LfD), we plan to develop a framework that will acquire and retain knowledge gathered from a teaching task, which draws inspiration from~\cite{agostini2020manipulation,caccavale2019kinesthetic}.
Actions at the {\sc{planfoon}} level can be connected to manipulation planning to define sequences of primitive actions (manipulation plan) to ground them. This sequence is generated using planning operators (POs) that encode physical cause-effects in terms of object-centered predicates.
Similar to~\cite{agostini2020manipulation}, we plan to use LfD to obtain DMP parameters that are associated to POs through action contexts.
However, as opposed to schemas~\cite{caccavale2019kinesthetic}, these POs will be connected to functional units at the the {\sc{planfoon}} level, which are then connected to descriptive knowledge at the {\sc{unifoon}} level.
Knowledge at these two levels of hierarchy can also be learned from demonstration.
For {\sc{planfoon}}-level learning, the main focus would be to identify the atomic skills needed for each {\sc{unifoon}} action.
A human demonstrator will teach the robot the necessary actions from start to end from which the robot will learn object features and parameters that are needed to identify the required items.
In the case where no {\sc{unifoon}} knowledge is available, then together with teaching skills, it is necessary to learn a {\sc{unifoon}}-level graph.
On one hand, this can be achieved via activity recognition; however, at the beginning, a simpler approach will be taken, where a human demonstrator will demonstrate an entire task to the robot while giving an explanation to the robot about its actions from which a graph may be constructed in a higher-level terminology. 
This graph can then be verified by the demonstrator to see if the robotic system has correctly created the {\sc{unifoon}} annotations.


\subsubsection{Example}
In Figure~\ref{fig:integration}, we illustrate an example of how a {\sc{unifoon}} functional unit for the pouring action may be decomposed into several simpler units at the {\sc{planfoon}} level.
Our understanding of the pour action could be translated into several sub-actions that are usually treated as atomic motion primitives from a robotics perspective.
Hence, an important starting point would be to first identify different motor skills that can be generalized to different object types or instances or that can be chained together for flexible execution.


\section{Conclusion and Future Work}
\label{sec:con}
In summary, in this preliminary work, we outline a road-map towards a robotic application of the functional object-oriented network (FOON) representation, which has not been extensively explored due to the domain-independent nature of FOON.
We propose to ground abstract knowledge in a regular FOON (which we refer to as the universal FOON or {\sc{unifoon}}) to a domain-specific level of FOON (which we refer to as a planetary FOON or {\sc{planfoon}}) using learning from demonstration (LfD).
Using LfD, a teacher can instruct a robot on how to execute high-level actions typically described in a universal FOON using its basic motion primitives while learning the necessary parameters for object detection and manipulation.
This framework aims to learn the object and motion parameters that are needed to physically ground abstract {\sc{unifoon}}-level concepts to the real world.
In the near future, we will further explore sub-components needed to realize this framework and evaluate our approach in both simulation and real-world settings.

\section*{Acknowledgement}
\noindent This research was funded by the Helmholtz Association.

\bibliographystyle{unsrt}
\bibliography{ref}

\end{document}